%% file: ijcai_paper.tex
\documentclass{article}
\usepackage{ijcai25}

\usepackage{times}
\usepackage{soul}
\usepackage{url}
\usepackage[hidelinks]{hyperref}
\usepackage[utf8]{inputenc}
\usepackage[small]{caption}
\usepackage{graphicx}
\usepackage{amsmath}
\usepackage{amsthm}
\usepackage{booktabs}

\usepackage{dsfont}
\usepackage{algorithm}
\usepackage{algorithmic}
\usepackage[switch]{lineno}
\usepackage[dvipsnames]{xcolor}

\def\gD{{\mathcal{D}}}
\def\gE{{\mathcal{E}}}
\def\gL{{\mathcal{L}}}
\def\gQ{{\mathcal{Q}}}
\def\gS{{\mathcal{S}}}
\def\mI{{\textit{I}}}
\def\mD{{\textit{D}}}

\title{DepthART: Monocular Depth Estimation as Autoregressive Refinement Task}

\author{
Bulat Gabdullin$^{1,2}$
\and
Nina Konovalova$^1$\and
Nikolay Patakin$^{1}$\and
Dmitry Senushkin$^{1}$\And
Anton Konushin$^{1}$\\
\affiliations
$^1$AIRI, Moscow, Russia\\
$^2$HSE University\\
\emails
\{gabdullin, konovalova, patakin, senushkin, konushin\}@airi.net\\
}
\begin{document}

\maketitle

\begin{abstract}
    Monocular depth estimation has seen significant advances through discriminative approaches, yet their performance remains constrained by the limitations of training datasets. While generative approaches have addressed this challenge by leveraging priors from internet-scale datasets, with recent studies showing state-of-the-art results using fine-tuned text-to-image diffusion models, there is still room for improvement. Notably, autoregressive generative approaches, particularly Visual AutoRegressive modeling, have demonstrated superior results compared to diffusion models in conditioned image synthesis, while offering faster inference times. 
In this work, we apply Visual Autoregressive Transformer (VAR) to the monocular depth estimation problem. However, the conventional GPT-2-style training procedure (teacher forcing) inherited by VAR yields suboptimal results for depth estimation. To address this limitation, we introduce DepthART - a novel training method formulated as a Depth Autoregressive Refinement Task. Unlike traditional VAR training with static inputs and targets, our method implements a dynamic target formulation based on model outputs, enabling self-refinement. By utilizing the model's own predictions as inputs instead of ground truth token maps during training, we frame the objective as residual minimization, effectively reducing the discrepancy between training and inference procedures.
Our experimental results demonstrate that the proposed training approach significantly enhances the performance of VAR in depth estimation tasks. When trained on Hypersim dataset using our approach, the model achieves superior results across multiple unseen benchmarks compared to existing generative and discriminative baselines.

\end{abstract}

\input{sections/intro}

\input{sections/related}

\input{sections/preliminary}
\input{sections/method}

\input{sections/experiments}
\input{sections/ablation}
\input{sections/discussion}
\input{sections/conclusion}
\section*{Contribution Statement}
Dmitry Senushkin - project leader
\bibliographystyle{named}
\bibliography{aaai25}

\end{document}

%% file: sections/intro.tex
\newcommand\blfootnote[1]{%
  \begingroup
  \renewcommand\thefootnote{}\footnote{#1}%
  \addtocounter{footnote}{-1}%
  \endgroup
}
\section{Introduction}
\label{sec:intro}

Monocular depth estimation (MDE) is a fundamental problem in computer vision. 
Depth maps provide a compact intermediate scene representation useful for decision making in physical surroundings. 
Recovering depth data from a single image promises a high practical value for different applications including spatial vision intelligence~\cite{wang2019pseudo}, autonomous driving~\cite{wang2019pseudo} and robotics~\cite{wofk2019fastdepth}. 

Early learning-based approaches~\cite{eigen2014} tackle the monocular depth estimation problem as a supervised regression task.
However, these methods were domain-specific \cite{silberman2012indoor} and heavily relied on annotated datasets. As a result, they were subject to a limited generalization ability caused by a low amount of annotated data available. Recent techniques suggested different tricks challenging this limitation. MiDaS ~\cite{ranftl2020towards} proposed to mitigate this issue by using an affine invariant depth training scheme on a mixture of datasets. While newer approaches proposing annotated data sources still appear~\cite{yang2024depth}, the acquisition of accurate depth annotations at scale remains challenging.


Recent studies~\cite{ke2024repurposing,fu2024geowizard} have highlighted the effectiveness of text-to-image diffusion models, originally trained on internet-scale image-caption datasets, as priors for monocular depth estimation. These approaches involve fine-tuning a pretrained diffusion model on a smaller, synthetic dataset with depth annotations, resulting in models that generate accurate and highly detailed depth maps. Concurrently, advancements in autoregressive models, such as the Visual AutoRegressive modeling ~\cite{tian2024visual} and LLaMA-Gen~\cite{sun2024autoregressive}, have demonstrated the capability of these models to generate high-quality images in class- or text-guided settings. These findings motivate an exploration of autoregressive generative techniques for depth estimation, offering a promising new direction.

In this work, we introduce\footnote{\url{https://bulatko.github.io/depthart-pp/}} a novel approach to monocular depth estimation based on the Visual AutoRegressive modeling~\cite{tian2024visual}. 
Our core contribution is the novel training procedure formulated as Depth Autoregressive Refinement Task. Our approach constructs dynamic targets using the model’s own predictions, rather than relying on ground truth token maps during training. By framing the objective as residual minimization and using model predictions as inputs, we bridge the gap between training and inference stages in autoregressive modeling, leading to enhanced depth estimation quality.
We validate our model extensively comparing it with popular baselines under similar conditions. 
To the best of our knowledge, this is the first autoregressive depth estimation model. Moreover, it performs on-par or superior compared with popular depth estimation baselines. 

Eventually, we formulate our contributions as follows:

\begin{enumerate}
    \item We introduce a novel application of Visual Autoregressive Transformer for depth estimation.
    \item We propose a new training paradigm for depth estimation, termed the Depth Autoregressive Refinement Task (DepthART), which facilitates self-refinement and bridges the gap between training and inference procedures.
    \item We demonstrate, through extensive experiments, that the Visual Autoregressive Transformer trained with DepthART achieves competitive or superior performance compared to existing baselines across several benchmarks not seen during training.
\end{enumerate}

%% file: sections/related.tex
\section{Related Work}
\label{sec:related}

\subsection{Monocular Depth Estimation}
Learning-based monocular depth estimation approaches can be broadly categorized into two main branches: metric and relative depth estimation methods.
Metric depth estimation~\cite{laina2016deeper,bhat2021adabins,yin2023metric3d} focuses on regressing absolute predictions at a metric scale. These models are typically trained on small, domain-specific datasets, which limits their ability to generalize efficiently across diverse environments. At the same time, relative depth estimation  methods aim to estimate depth up to unknown shift and scale (SSI) or just unknown scale (SI). MiDaS ~\cite{ranftl2020towards} introduced shift and scale invariant depth training on a mixture of several domain-specific datasets, significantly improving model generalization. Despite it, the depth predictions remained geometrically incomplete, i.e. point clouds cannot be built using model predictions. GP2 ~\cite{patakin2022single} addressed this limitation by proposing an end-to-end training scheme that estimates a scale-invariant, geometry-preserving depth maps. Meanwhile, two-stage pipelines were developed to reduce shift ambiguity in depth maps at the second stage~\cite{yin2021learning} or to upgrade the depth map to metric scale~\cite{bhat2023zoedepth}. Further advancements in the field have been driven by the integration of various priors~\cite{yang2024depth,yin2023metric3d}, improvements in architectural designs~\cite{ranftl2021vision,agarwal2023attention,ning2023trap} and the expansion of training data~\cite{yang2024depth}.
 
\subsection{Generative Modeling}
Recently, diffusion models have demonstrated their versatility across various computer vision tasks, including image generation ~\cite{ho2020denoising,rombach2022high}, video generation ~\cite{blattmann2023stable}, or 3D objects modeling ~\cite{melas2023realfusion}. Beyond these applications, diffusion models have also been successfully employed in other problems, such as depth estimation \cite{duan2023diffusiondepth}, image segmentation \cite{wang2023dformer} and object detection \cite{chen2023diffusiondet}. Notably, Marigold ~\cite{ke2024repurposing} and GeoWizard ~\cite{fu2024geowizard} have demonstrated that the Stable Diffusion model \cite{rombach2022high}, pretrained on the large-scale image-caption dataset LAION-5B \cite{schuhmann2022laion}, can produce high-quality depth maps after minor finetuning. This highlights the potential of utilizing pretrained generative models to enhance depth estimation accuracy and robustness across different domains.

\subsection{Autoregressive Modeling}
While diffusion models remain one of the most widely-used generative approach, recent advancements in autoregressive models have shown significant promise for various generative tasks \cite{yu2023language,sun2024autoregressive,tian2024visual}. These methods rely on discrete token-based image representations typically generated by VQ-VAE~\cite{van2017neural} or its derivatives. These derivatives often include architectural enhancements \cite{yu2021vector}, additional masking techniques \cite{huang2023not}, or the incorporation of adversarial and perceptual losses \cite{esser2021taming}. Autoregressive image synthesis is generally formulated as the sequential generation of tokens~\cite{esser2021taming}, followed by decoding them from the VQ space.
Many approaches employ the GPT-2 \cite{radford2019language} decoder-only architecture to predict sequences of VQ-VAE tokens \cite{esser2021taming,chang2023muse}. However recent works \cite{tian2024visual} have introduced the concept of predicting multi-scale token maps rather than token sequences. 
This approach reduces the risk of structural degradation and decreases the generation time for high-resolution images, enabling high-quality class- and text-conditioned image generation.



%% file: sections/preliminary.tex
\section{Preliminary}
\label{sec:preliminary}

\begin{figure}[t]
\centering
\includegraphics[clip, trim=0.6cm 0.4cm 0.6cm 0.35cm, width=1.0\columnwidth]{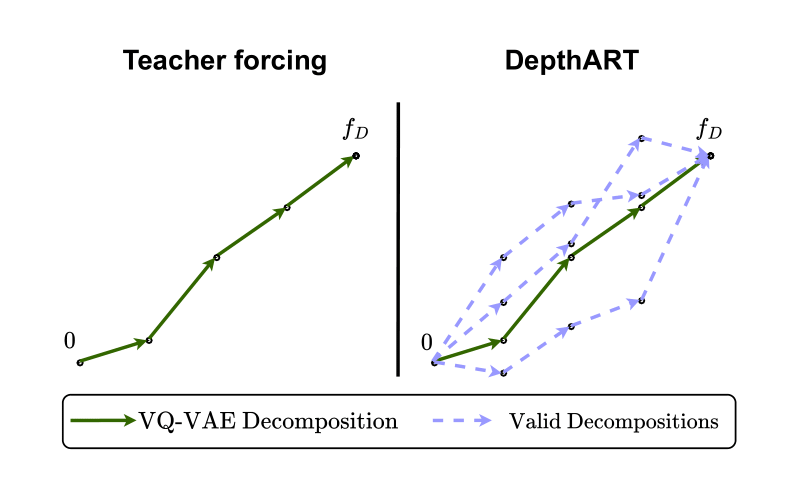}
\caption{VQ-VAE provides a single decomposition, while there are different trajectories that can result in the same features. The usage of teacher forcing during training leads to selecting only one trajectory, given by the VQ-VAE decomposition. In contrast, DepthART enables the use of other possible decompositions, offering diverse refinement paths compared to the unique guidance used in teacher forcing.}
\label{fig:multimodal}
\end{figure}


\begin{figure*}[!ht]
\centering
\includegraphics[clip, trim=.6cm 0.0cm 0.6cm 0.2cm, width=1.\linewidth]{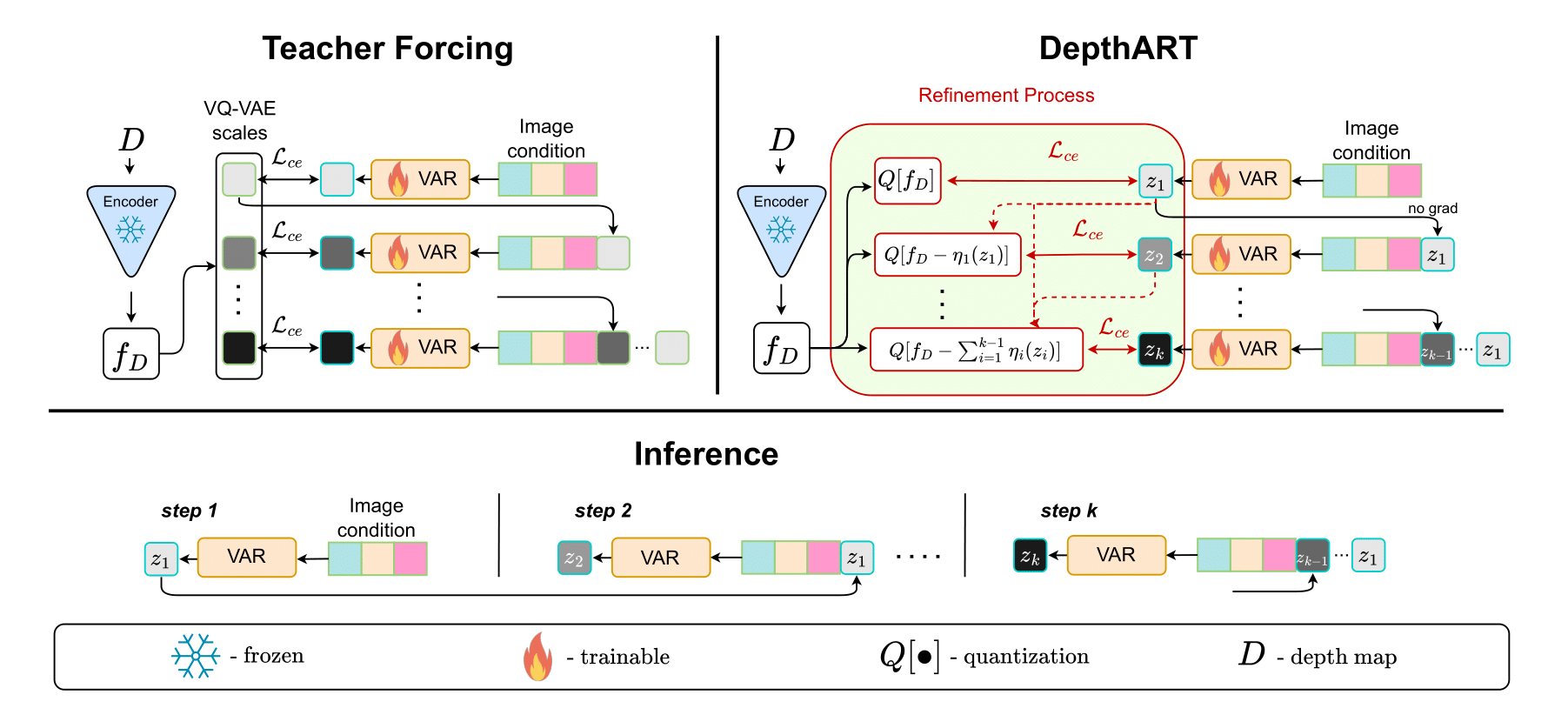} 
\caption{
DepthART (right). We also demonstrate how our training procedure better aligns with the inference process (shown at the bottom), as in both training and inference, the model operates on its own predictions as inputs. In the teacher forcing approach, quantized token maps provided by VQ-VAE serve as both inputs and targets during training. Our DepthART method introduces a refinement process (highlighted in the red box), where the model self-refines by using its predicted token maps as inputs instead of predefined VQ-VAE scales. The targets are defined as the quantized residuals between the encoded depth features $f_D$ and the cumulative model predictions up to the current scale. Depth features $f_D$ are extracted from the VQ-VAE encoder without undergoing quantization.} 

\label{fig:pipeline}
\end{figure*}

\paragraph{Next-scale visual autoregressive modeling.} Typically autoregressive image generation involves predicting image tokens in a raster scan order.
However, recent work~\cite{tian2024visual} introduced a novel autoregressive training approach for class-conditional image generation - visual autoregressive modeling. Instead of predicting tokens individually, they proposed generating token maps with varying scale. Each predicted token map progressively increases in resolution compared to the previous one, resulting in a scale-wise decomposition of the image. 
Specifically, an image $\mI$ is modelled as a sequence of $\{x_1, x_2, \dots, x_K\}$, where $x_k$ is a token map of size $s_k = (h_k, w_k)$. 
The visual autoregressive transformer based on GPT-2 is then trained to maximize a class-conditioned likelihood:
\begin{equation}\label{scale:ar}
    p(x_1, x_2, \dots, x_K | c) = \prod_{k=1}^K p(x_k|x_1, x_2, \dots, x_{k-1}, c)
\end{equation}

\noindent Unlike the traditional token-by-token prediction approach, which can cause structural degradation due to raster scan ordering, visual autoregressive modeling predicts images scale-by-scale. Given that depth maps, like images, exhibit high spatial correlations, we explore the applicability of visual autoregressive modeling for solving depth estimation problem.

\paragraph{Discrete image representations.} Effective autoregressive image modeling relies on discretizing images into a finite set of tokens, a task closely related to image compression. Image compression methods have evolved from linear projections to advanced neural approaches, such as vector quantized variational autoencoders (VQ-VAE)~\cite{van2017neural}. VQ-VAE approach combines an encoder-decoder neural network with a learnable token codebook of finite size. The encoder compresses the input image into a latent space, which is then quantized by mapping the latent vectors to the closest entries in the codebook. This quantized representation is then decoded back into the image space, with the entire process trained end-to-end using a log-likelihood objective over the image distribution.

Visual autoregressive model training requires multi-scale image decomposition represented as a sequence of token maps. Differently from original VQ-VAE~\cite{van2017neural} an input image $\mI$ is quantized by encoder $\gE$ into $K$ token maps $\{ x_1, x_2, \dots, x_K \}$ with resolutions $\{(h_k, w_k)\}_{k=1,K}$. Quantization operation $Q[\cdot]$ is performed using the same codebook $Z$ regardless of scale. Combined using upsample-convolution operators $\eta_i$ token maps are assumed to sum up into continuous features $\gE(\mI)$. Accordingly, the $k$-th scale map is calculated as a scaled and quantized residual of the extracted features:
\begin{align}
    & r_k = \gE(\mI) - \sum_{i=1}^{k-1}\eta_i (x_i) \label{eq:residuals}\\
    & x_k = \gQ\big[\gS(r_k, s_k)\big] \label{eq:quantize}
\end{align}

\vspace{0.2cm}
$\gS$ - scaling operation.

Eventually, decoder $\gD$ recovers reconstructed image $\hat{\mI}$ from  given token maps:
\begin{equation}
    \hat{\mI} = \gD\bigg(\sum_{k=1}^{K}\eta_k (x_k)\bigg)
\end{equation}

In our approach, we adapt a pretrained modification of VQ-VAE tailored specifically for the visual autoregressive modeling transformer (VAR) \cite{tian2024visual}. While originally designed for colored images, we observe that VQ-VAE~\cite{tian2024visual} can be applied to encode depth maps as well.

%% file: sections/method.tex
\section{Method}\label{sec:method}

In this work, we formulate the monocular depth estimation task as an image-conditioned autoregressive generation problem. We leverage the Visual Autoregressive Transformer (VAR) for monocular depth estimation, adapting it to generate depth token maps from input images. As our primary contribution, we introduce the novel training procedure formulated as Depth Autoregressive Refinement Task -- DepthART. 


\begin{figure}[t]
\centering
\includegraphics[clip, trim=0.1cm 0.2cm 0.2cm 0.15cm, width=1.0\columnwidth]{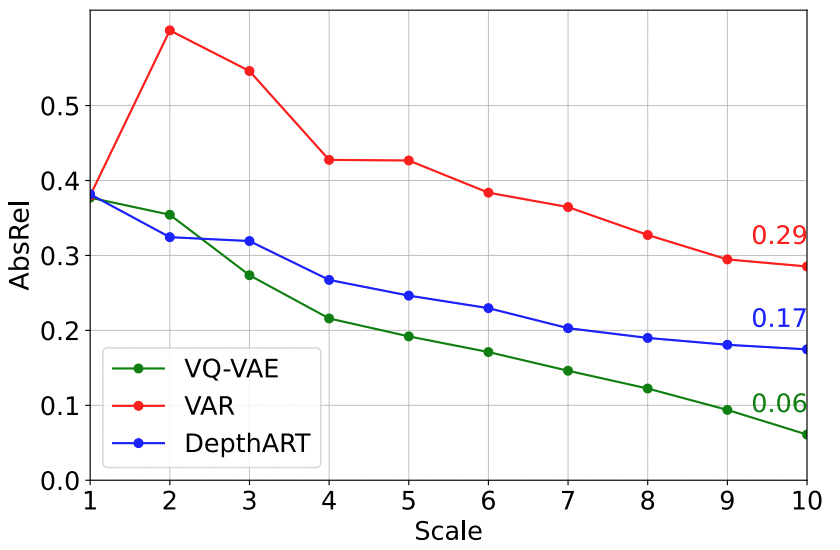}
\caption{VAR trained with DepthART demonstrates superior reconstruction quality at every
scale compared to the teacher forcing training procedure, achieving a \textbf{1.7x} improvement in reconstruction quality. Since we
do not finetune the VQ-VAE, its end-to-end reconstruction quality is shown as a soft limitation on reconstruction error.
}
\label{fig:impact}
\end{figure}


\subsubsection{Depth Autoregressive Refinement Task}
The original Visual AutoRegressive modeling~\cite{tian2024visual} relies on scale-wise decomposition of image provided by pre-trained VQ-VAE encoder. 
During training the model predicts next-scale token map from the previous sequence of ground truth token maps. Ground truth token maps are computed using VQ-VAE (Eq. \ref{eq:quantize}, \ref{eq:residuals}). The guidance objective is a cross-entropy loss between VAR prediction and the same scale token map provided by VQ-VAE. Inputting ground-truth token maps during training creates a discrepancy with inference, as the model never learns to account for its own prediction errors from previous steps.
However, since the final feature map is constructed as a sum of all scale contributions, and this sum can be achieved through different combinations of addends, we can reformulate the problem: at each step, we can compute the difference between the current accumulated sum and the target feature map, allowing us to define the target for each scale as a residual that needs to be predicted to reach the final feature value.

We address this issue and reformulate training objective as a \textbf{D}epth \textbf{A}utoregressive \textbf{R}efinement \textbf{T}ask (DepthART). Our main goal is to enable model self-refinement during training. Hence, we construct inputs and targets dynamically from model predictions. 
Let's consider an input image $\mI$ with corresponding ground truth-depth map $\mD$. We firstly encode an image into a series of token maps $\{x_1, x_2, \dots x_K\}$ Eq. \ref{eq:quantize} provided by VQ-VAE~\cite{tian2024visual}. Resulting image token maps are fed to the model input as a starting sequence and serve as a conditioning for depth map estimation. 
Constructing dynamic supervision targets in our approach starts with performing model inference for given image token maps. We denote predicted depth token maps as $\{z_1, z_2, \dots, z_K\}$:
\begin{equation}\label{eq:inference}
   z_k = \text{VAR}(z_1, \dots, z_{k-1}, x_1, \dots x_K)
\end{equation}
Next, we encode ground-truth depth $\mD$ with the same VQ-VAE encoder into continuous features $f_{\mD}$, discarding quantization process.
Residual prediction targets $\{t_1, t_2, \dots, t_K\}$ can be constructed based on encoded depth features $f_\mD$ and a series of models predictions up to current scale. 
This process is done similarly to VQ-VAE decomposition method (Eq. \ref{eq:residuals}, \ref{eq:quantize}):
\begin{align}
    \delta_k = f_{\mD} - \sum_{i=1}^{k-1}\eta_i(z_i) \\
    t_k = \gQ\big[ \gS(\delta_k, s_k) \big]
\end{align}

\vspace{0.2cm}

Eventually, the training objective takes the form of cross-entropy loss between predicted and target token maps:
\begin{equation}
    \gL = \sum_{k=1}^K \gL_{CE}(z_k, t_k)
\end{equation}
As the result we form a new set of training samples, where $z_k$ and $t_k$ are model inputs and targets respectively. 
In contrast to original VAR training, these token maps are dynamically constructed at every training step rather than relying on a single predefined VQ-VAE decomposition~(see Figure \ref{fig:pipeline}).

The training process benefits from such formulation in a few ways. Firstly, our procedure enables model self-refinement by making model aware of own predictions and framing the task as residual refinement. Since VQ-VAE~\cite{tian2024visual} decomposition comes from approximating continuous features with summation of discrete token maps, multiple plausible decompositions can exist. Secondly, we argue that exploring various ways of decomposing input into token maps is beneficial for model training. Thirdly, this training procedure better aligns with the inference process, where the model receives its own outputs at each step, thus reducing the training-inference discrepancy.

%% file: sections/experiments.tex
\begin{table*}[h!]

\centering
\renewcommand{\arraystretch}{1.5} 
\setlength{\tabcolsep}{8pt} 
\resizebox{\linewidth}{!}{


\begin{tabular}{l|c|c|c|cc|cc|cc|cccc|c}
\hline
\textbf{Models} & FLOPS & Train./Infer., \textit{ms} & \#Params  &  \multicolumn{2}{c|}{\textbf{ETH}} & \multicolumn{2}{c|}{\textbf{TUM}} & \multicolumn{2}{c|}{\textbf{NYU}} & \multicolumn{4}{c|}{\textbf{IBIMS}} & \textbf{Rank$\downarrow$} \\ \cline{5-14}
                        & & & & $\delta_1{\downarrow}$  & AbsRel${\downarrow}$  & $\delta_1{\downarrow}$  & AbsRel${\downarrow}$    & $\delta_1{\downarrow}$  & AbsRel${\downarrow}$   & $\delta_1{\downarrow}$  & AbsRel${\downarrow}$ &  
                        pe-fla${\downarrow}$ & 
                        pe-ori${\downarrow}$ &\\ \hline
\textbf{GP-2} (EffNet-B5) & 13.4G & 41.4 / 21.4 & 30M   &    0.23  &  0.175 & 0.427  &  0.247 & 0.162 & 0.125 & 0.162 & 0.121  & 4.70 & 12.8 & 5.2 \\
\textbf{Midas} (ResNeXt-101) & 92G & 28.1 / 10.9 & 104.5M  &  0.203 & \underline{0.160}  & 0.325 & 0.207 & 0.143  & 0.116  & 0.140  & 0.112  & 2.71 & 10.8 & 3.5 \\
\textbf{AdaBins} (EffNet-B5) & 317G & 26.8 / 13.4 & 78.2M  &  0.235 & 0.184 & 0.323 & 0.206  & 0.141  & 0.115 & 0.161 & 0.125 & 4.65 & 12.6 & 4 \\
\textbf{DiT-depth} (DiT) & 3.2T & 514.5 / 217.5 & 675M  &  0.309  & 0.220 & \textbf{0.252}  & \textbf{0.169} & 0.149  & 0.120  & 0.169 & 0.127  & 2.86 & 8.71 & 4.3  \\ 
\textbf{DPT} (ViT-L) & 963G & 128.7 / 47.6 & 344M  & \underline{0.198} & \textbf{0.150}  & 0.435  & 0.251 & \textbf{0.121} & \textbf{0.107} & \underline{0.134} & \underline{0.108} & 2.97 & \underline{8.14} & \underline{2.9}\\ 
\textbf{VAR} (GPT-2) & 1.17T & 187.9 / 74.8 & 310M & 0.245 & 0.285  & 0.396  & 0.294 & 0.185 & 0.141 & 
0.177 & 0.133 & \underline{1.98} & 9.44 & 6.1\\ 
\hline
\textbf{VAR+DepthART} (Ours, GPT-2) & 1.17T & 192.2 / 74.8 & 310M  & \textbf{0.196} & 0.177 & \underline{0.275}  & \underline{0.178} & \underline{0.141}  & \underline{0.115} & \textbf{0.129} & \textbf{0.106} & \textbf{1.91} & \textbf{7.27} & \textbf{2.1} \\
\end{tabular}
}


\caption{Quantitative evaluation across benchmarks unseen during training. Overall performance is summarized using a rank metric. VAR trained with DepthART outperforms the original VAR trained with teacher forcing and achieves the highest overall performance among a diverse set of depth estimation baselines. Trainable parameters count, FLOPS, Training iteration time and Inference time on 1×H100 with batch size 1 are also provided.}
\label{tab:model_comparison_sil2_extended}
\end{table*}

\begin{figure*}[t]
\centering
\includegraphics[width=0.97\textwidth]{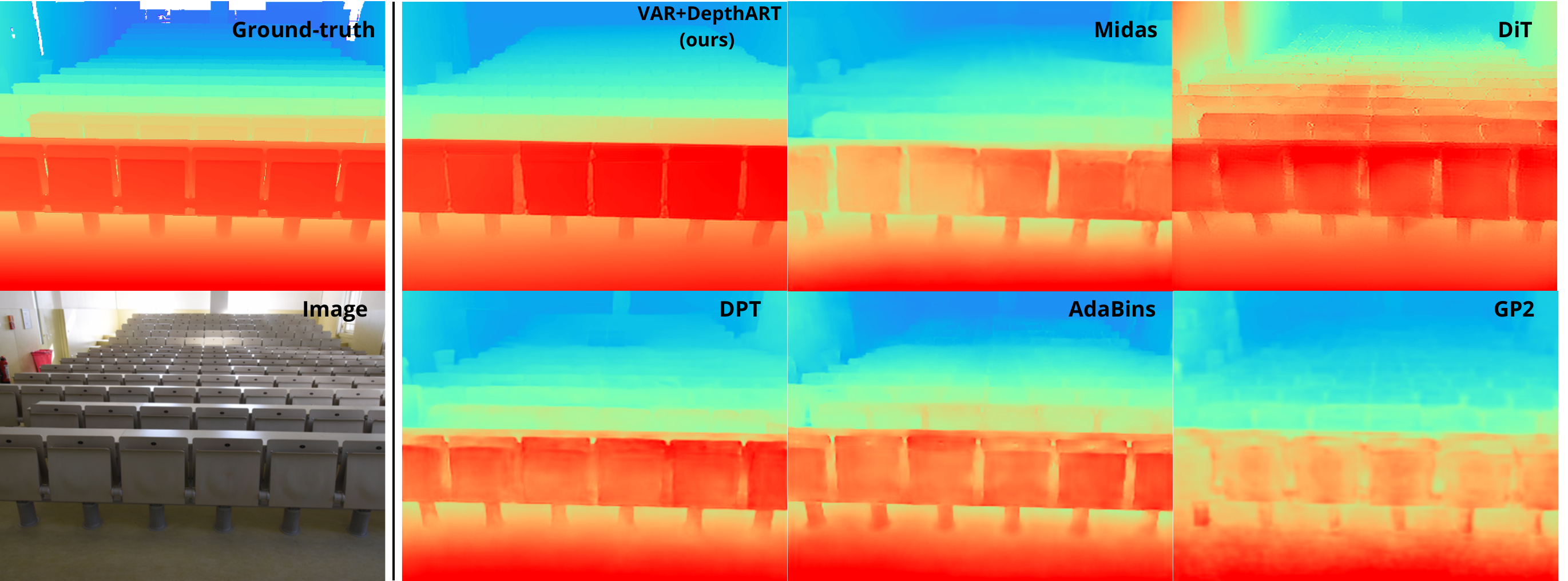}
\includegraphics[width=0.97\textwidth]{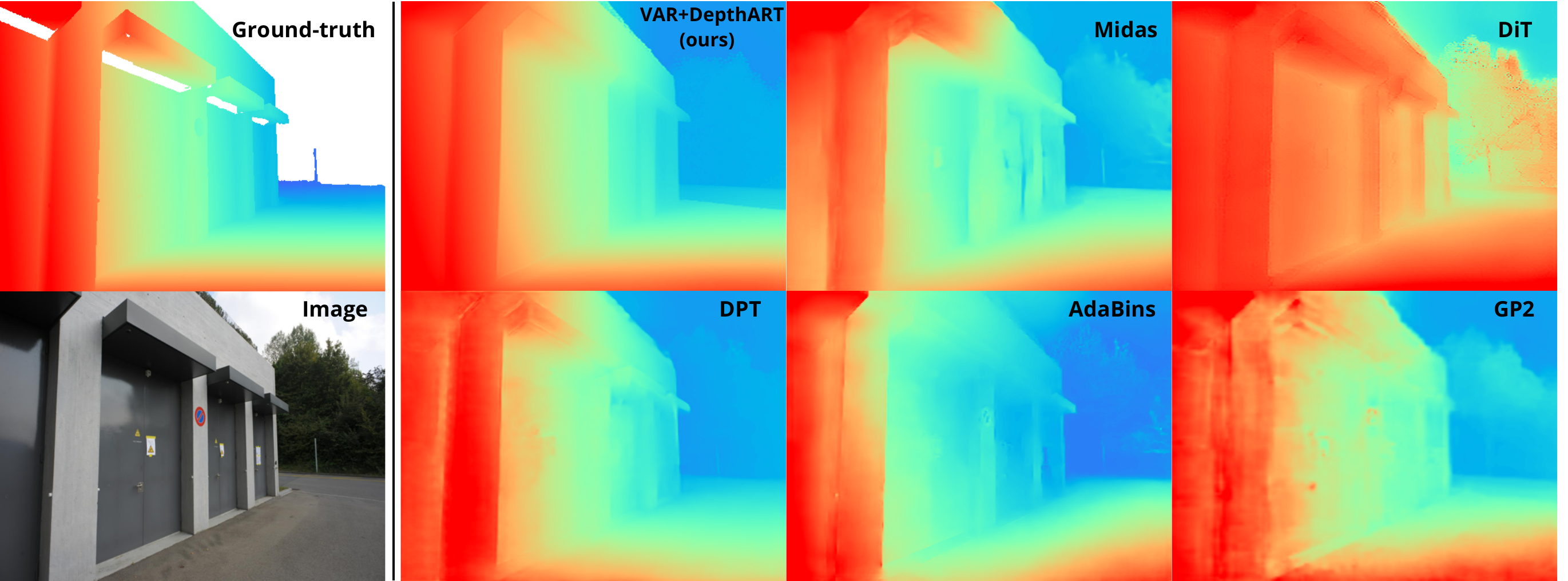} 
\caption{Qualitative comparison of the VAR trained with DepthART against various baselines. Our model generates more precise depth estimates in planar regions while maintaining the overall scene structure.}
\label{fig:depthmaps}
\end{figure*}

\begin{figure*}[!ht]
\centering
\includegraphics[width=\textwidth]{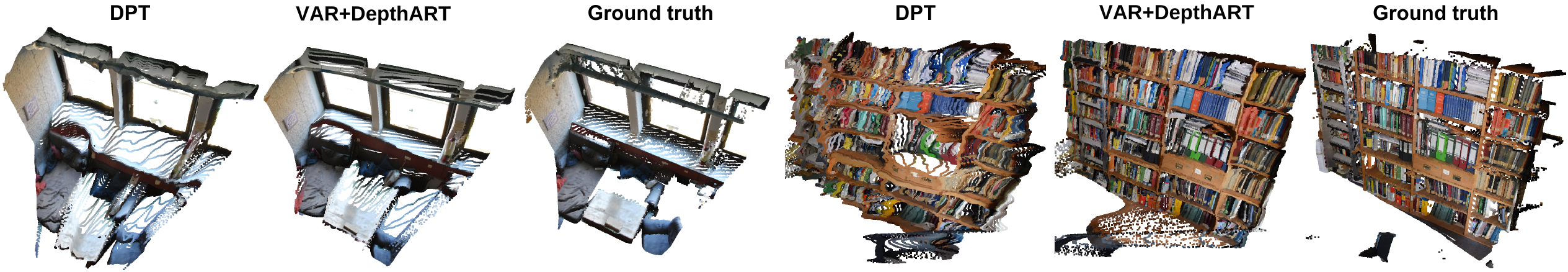}
\caption{Qualitative comparison of point clouds reconstructed from predicted depth maps on the IBIMS dataset. The visual autoregressive transformer trained with DepthART delivers higher-quality reconstructions, particularly in planar regions.}\label{fig:pointclouds}
\end{figure*}

\section{Experiments}
\label{sec:experiments}


In this section, we empirically validate the effectiveness of our proposed approach on several depth estimation benchmarks that were not used during training. Specifically, we demonstrate that: (1) the usage of DepthART training method significantly enhances the performance of the Visual Autoregressive Transformer applied to depth estimation task compared to the usage of teacher forcing, (2) the Visual Autoregressive Transformer trained with DepthART achieves comparable or superior accuracy relative to other baselines. 

\subsubsection{Implementation Details}
For our experimental evaluation, we chose to predict depth maps up to an unknown scale (SI). Unlike the commonly used scale-and-shift invariant training approach~\cite{ranftl2020towards}, scale-invariant training preserves the ability to reconstruct geometry from the predicted depth maps, which is essential for practical applications like single-view reconstruction. Therefore, prior to inputting the depth maps into VQ-VAE, we apply the following transformation:
\begin{equation}\label{eq:SI}
    \mD_{norm} = \frac{\mD}{\mD_{98} + \epsilon} \times 2 - 1
\end{equation}
where $\mD_{98}$ -- $98\%$ percentile of individual depth map. 

Visual Autoregressive Transformer is trained with DepthART using AdamW~\cite{loshchilov2018decoupled} optimizer with a learning rate of $10^{-4}$ and weight decay of $10^{-2}$ and batch size equals to 4. Additionally we decrease learning rate during training with StepLR scheduler with a step size of $10,000$ and a gamma of $0.8$. Training of our model takes 17 hours using 4 NVIDIA H100 GPUs. 





\paragraph{Training protocol.} To ensure consistent training conditions across all models, we train both the Visual Autoregressive Transformer and baseline models on the same dataset. Due to the requirement of dense ground-truth depth maps for variational autoencoders, we utilize the highly realistic synthetic HyperSim dataset~\cite{roberts2021hypersim}, which includes 461 diverse indoor scenes. The pretrained VQ-VAE~\cite{tian2024visual} used in our experiments generates multi-scale token maps only up to a maximum resolution of 256$\times$256, so we train all models at this resolution. 



\paragraph{Evaluation protocol.} Evaluation is performed on four datasets unseen during training: NYUv2~\cite{silberman2012indoor} and IBIMS ~\cite{ibims} capturing indoor environments, TUM ~\cite{li2019learning} capturing dynamic humans in indoor environment, ETH3D~\cite{schops2017multi} providing high-quality depth maps for outdoor environments. 
\noindent Since all models trained to predict depth maps up to unknown scale, we first align predictions with ground-truth depth maps in terms of $\mathcal{L}_1$. Firstly, we evaluate accuracy of estimated depth maps using two commonly used metrics: Absolute Mean Relative Error (AbsRel) ($\downarrow$) and $\delta_1(\downarrow)$. Additionally, we assess the predicted depth maps using  depth planar region deviations (pe-fla $\downarrow$) and plane orientation error (pe-ori, in $^{\circ}, \downarrow$) on IBIMS dataset \cite{ibims}. Recent state-of-the-art generative depth estimation methods have shown that large-scale pretraining plays a crucial role in reducing the synthetic-to-real domain gap. For instance, popular generative models like Marigold ~\cite{ke2024repurposing} leverage internet-scale pretraining on images before being fine-tuned on small synthetic datasets for depth estimation. This strong image prior significantly decreases the synthetic-to-real domain gap, enabling these generative models to achieve high zero-shot performance on real-world depth estimation benchmarks. Therefore, we do not perform additional steps to address the domain gap. 

\paragraph{Baselines.} We evaluate our approach against a diverse set of baseline models, organized into three categories. First, we consider several widely used depth estimation architectures trained discriminatively with an $\mathcal{L}_2$ regression loss, including \textbf{MiDaS}~\cite{ranftl2020towards}, \textbf{GP2}~\cite{patakin2022single}, and \textbf{DPT}~\cite{ranftl2021vision}. We also include \textbf{AdaBins}~\cite{bhat2021adabins}, which represents a classification-based approach to depth estimation. Additionally, we evaluate \textbf{DiT}~\cite{peebles2023scalable}, a transformer-based diffusion model pretrained on ImageNet. Finally, to assess the impact of our training procedure, we compare VAR trained with DepthART (\textbf{VAR+DepthART}) against the \textbf{VAR} trained with original procedure - teacher forcing. More baseline training details are provided in the supplementary materials.

\subsection{Experimental Results}

\subsubsection{Teacher Forcing vs DepthART Training}

To demonstrate the advantages of our approach, we trained VAR on the HyperSim dataset using both the original teacher forcing training procedure and our DepthART method. Figure \ref{fig:impact} presents a detailed comparison of these training methods, evaluated on the ETH3D dataset.
We assess the reconstruction quality by calculating the AbsRel metric (Figure \ref{fig:impact}) between the intermediate depth maps decoded at each autoregression step, based on cumulative predictions from both teacher forcing and DepthART. Since we used a pretrained VQ-VAE without fine-tuning it for depth estimation, we also present its end-to-end depth map reconstruction error as a soft limit achievable by VAR.
While the model trained using the teacher forcing approach struggles to improve reconstruction quality at early scales, the DepthART-trained model consistently refines its predictions, achieving an overall reduction in relative error of approximately 70\% compared to the baseline. Notably, the DepthART-trained model discovers a slightly better decomposition at the second scale than the reference provided by VQ-VAE, highlighting the non-uniqueness and potential suboptimality of the VQ-VAE decomposition.

\subsubsection{Comparison Against Baselines}

To prove the efficiency of autoregressive approach in depth estimation, we train a set of popular baselines in similar conditions. Table \ref{tab:model_comparison_sil2_extended} presents the evaluation  results of all models on benchmarks that were not seen during training. To assess overall performance, we calculated each model's rank on every dataset, and then averaged these ranks across all datasets. As can be seen from the Table \ref{tab:model_comparison_sil2_extended}, the VAR trained with DepthART achieves the best overall performance. Notably, both VAR models trained with the teacher forcing and DepthART methods showed significantly better planar depth accuracy on the IBIMS dataset. We provide qualitative comparison of predicted depth maps in Figure \ref{fig:depthmaps}. Besides, point clouds reconstructed from predicted depth maps (Figure \ref{fig:pointclouds}) further support this observation. These results highlight the potential of autoregressive models for depth estimation.

%% file: sections/ablation.tex
\section{Ablation Studies}
In this section, we analyze the impact of generative pretraining. Since we employ Visual Autoregressive Transformer, a generative model, for depth estimation, the generative pretraining stage plays a crucial role in the model's performance. Understanding and effectively operating in the VQ-VAE latent space is a challenging task, and the Visual Autoregressive Transformer learns to navigate this space during the pretraining phase. To evaluate the importance of pretraining, we conducted an experiment without the class-conditioned generation pretraining on ImageNet. 

As shown in Table~\ref{tab:pretrain_comparison}, the absence of pretraining leads to significantly degraded metrics across all evaluation criteria. These results demonstrate that the knowledge acquired during generative pretraining is essential for effective depth estimation. Furthermore, we hypothesize that utilizing more powerful pretraining schemes, particularly text-to-image generation on large-scale datasets like LAION-5B~\cite{schuhmann2022laion}, could yield even better performance.

\begin{table}[!t]
\centering
\renewcommand{\arraystretch}{1.4}
\setlength{\tabcolsep}{10pt}
\resizebox{\linewidth}{!}{

\begin{tabular}{l|cc|cc|cc}
\hline
 & \multicolumn{2}{c|}{\textbf{ETH}} & \multicolumn{2}{c|}{\textbf{TUM}} & \multicolumn{2}{c}{\textbf{IBIMS}} \\ \cline{2-7} 
                           & $\delta_1 \downarrow$  & AbsRel$\downarrow$ & $\delta_1 \downarrow$ & AbsRel$\downarrow$ & $\delta_1 \downarrow$ & AbsRel$\downarrow$ \\ \hline
w/o ImageNet pretrain        &   0.392 & 0.46 & 0.561 & 0.522 & 0.338 & 0.301 \\
with ImageNet pretrain        &   \textbf{0.196}     &  \textbf{0.177}       &   \textbf{0.275}     &    \textbf{0.178}  & \textbf{0.129} & \textbf{0.106}  \\ \hline
\end{tabular}
}
\caption{Ablation results demonstrate that the generative pretraining of Visual Autoregressive Transformer is crucial for model performance, with metrics showing significant degradation when pretraining is omitted.}
\label{tab:pretrain_comparison}
\end{table}

%% file: sections/discussion.tex
\section{Discussion}
\label{sec:discussion}


This work demonstrates the potential of generative autoregressive modeling for monocular depth estimation. Currently, our visual autoregressive transformer builds on pretraining, which is constrained by the the ImageNet dataset. We believe that pretraining on more extensive and diverse datasets, such as those used for text-to-image generation, could significantly enhance our model's performance. A primary limitation of our approach is the reliance on components from the original VAR: both the VQ-VAE network and the generative pretraining of the transformer. The VQ-VAE was trained at low resolution on the relatively smaller OpenImages dataset~\cite{kuznetsova2020open}, in contrast to larger, more recent datasets like LAION-5B~\cite{schuhmann2022laion}. Another potential limitation stems from the VQ-VAE training specifics: there is no guarantee that the quantization operation can accurately represent any scale that serves as a target during training. This limitation may lead to scenarios where the model converges to a local minimum due to the inherent constraints of the VQ-VAE's learned codebook. We anticipate that upgrading to a higher-quality VQ-VAE and employing stronger pretraining could greatly benefit our method, and we identify these limitations as key directions for future research.



%% file: sections/conclusion.tex
\section{Conclusion}
\label{sec:conclusion}

In this paper, we tackle the depth estimation problem through an autoregressive lens, specifically adapting the visual autoregressive modeling approach~\cite{tian2024visual} for this task. Originally designed for class-conditioned image generation, we repurposed the Visual Autoregressive Transformer for image-conditioned depth map estimation. Our analysis highlights limitations in the standard VAR training process - teacher forcing, which leads to suboptimal accuracy on public depth benchmarks. To address these challenges, we proposed a novel training formulation, the Depth Autoregressive Refinement Task (DepthART). The Visual Autoregressive Transformer trained with DepthART showed substantial performance improvements over the teacher forcing procedure and achieved competitive or superior results on public benchmarks compared to recent methods. Our approach enhances the model’s self-refinement ability and resolves the the training-inference discrepancy  issues of visual autoregressive modeling, as demonstrated through empirical evaluation.

%% file: ijcai_paper.bbl
\begin{thebibliography}{}

\bibitem[\protect\citeauthoryear{Agarwal and Arora}{2023}]{agarwal2023attention}
Ashutosh Agarwal and Chetan Arora.
\newblock Attention attention everywhere: Monocular depth prediction with skip attention.
\newblock In {\em Proceedings of the IEEE/CVF Winter Conference on Applications of Computer Vision}, pages 5861--5870, 2023.

\bibitem[\protect\citeauthoryear{Bhat \bgroup \em et al.\egroup }{2021}]{bhat2021adabins}
Shariq~Farooq Bhat, Ibraheem Alhashim, and Peter Wonka.
\newblock Adabins: Depth estimation using adaptive bins.
\newblock In {\em Proceedings of the IEEE/CVF conference on computer vision and pattern recognition}, pages 4009--4018, 2021.

\bibitem[\protect\citeauthoryear{Bhat \bgroup \em et al.\egroup }{2023}]{bhat2023zoedepth}
Shariq~Farooq Bhat, Reiner Birkl, Diana Wofk, Peter Wonka, and Matthias M{\"u}ller.
\newblock Zoedepth: Zero-shot transfer by combining relative and metric depth.
\newblock {\em arXiv preprint arXiv:2302.12288}, 2023.

\bibitem[\protect\citeauthoryear{Blattmann \bgroup \em et al.\egroup }{2023}]{blattmann2023stable}
Andreas Blattmann, Tim Dockhorn, Sumith Kulal, Daniel Mendelevitch, Maciej Kilian, Dominik Lorenz, Yam Levi, Zion English, Vikram Voleti, Adam Letts, et~al.
\newblock Stable video diffusion: Scaling latent video diffusion models to large datasets.
\newblock {\em arXiv preprint arXiv:2311.15127}, 2023.

\bibitem[\protect\citeauthoryear{Chang \bgroup \em et al.\egroup }{2023}]{chang2023muse}
Huiwen Chang, Han Zhang, Jarred Barber, AJ~Maschinot, Jose Lezama, Lu~Jiang, Ming-Hsuan Yang, Kevin Murphy, William~T Freeman, Michael Rubinstein, et~al.
\newblock Muse: Text-to-image generation via masked generative transformers.
\newblock {\em arXiv preprint arXiv:2301.00704}, 2023.

\bibitem[\protect\citeauthoryear{Chen \bgroup \em et al.\egroup }{2023}]{chen2023diffusiondet}
Shoufa Chen, Peize Sun, Yibing Song, and Ping Luo.
\newblock Diffusiondet: Diffusion model for object detection.
\newblock In {\em Proceedings of the IEEE/CVF international conference on computer vision}, pages 19830--19843, 2023.

\bibitem[\protect\citeauthoryear{Duan \bgroup \em et al.\egroup }{2023}]{duan2023diffusiondepth}
Yiqun Duan, Xianda Guo, and Zheng Zhu.
\newblock Diffusiondepth: Diffusion denoising approach for monocular depth estimation.
\newblock {\em arXiv preprint arXiv:2303.05021}, 2023.

\bibitem[\protect\citeauthoryear{Eigen \bgroup \em et al.\egroup }{2014}]{eigen2014}
David Eigen, Christian Puhrsch, and Rob Fergus.
\newblock Depth map prediction from a single image using a multi-scale deep network.
\newblock In Z.~Ghahramani, M.~Welling, C.~Cortes, N.~Lawrence, and K.Q. Weinberger, editors, {\em Advances in Neural Information Processing Systems}, volume~27. Curran Associates, Inc., 2014.

\bibitem[\protect\citeauthoryear{Esser \bgroup \em et al.\egroup }{2021}]{esser2021taming}
Patrick Esser, Robin Rombach, and Bjorn Ommer.
\newblock Taming transformers for high-resolution image synthesis.
\newblock In {\em Proceedings of the IEEE/CVF conference on computer vision and pattern recognition}, pages 12873--12883, 2021.

\bibitem[\protect\citeauthoryear{Fu \bgroup \em et al.\egroup }{2024}]{fu2024geowizard}
Xiao Fu, Wei Yin, Mu~Hu, Kaixuan Wang, Yuexin Ma, Ping Tan, Shaojie Shen, Dahua Lin, and Xiaoxiao Long.
\newblock Geowizard: Unleashing the diffusion priors for 3d geometry estimation from a single image.
\newblock {\em arXiv preprint arXiv:2403.12013}, 2024.

\bibitem[\protect\citeauthoryear{Ho \bgroup \em et al.\egroup }{2020}]{ho2020denoising}
Jonathan Ho, Ajay Jain, and Pieter Abbeel.
\newblock Denoising diffusion probabilistic models.
\newblock {\em Advances in neural information processing systems}, 33:6840--6851, 2020.

\bibitem[\protect\citeauthoryear{Huang \bgroup \em et al.\egroup }{2023}]{huang2023not}
Mengqi Huang, Zhendong Mao, Quan Wang, and Yongdong Zhang.
\newblock Not all image regions matter: Masked vector quantization for autoregressive image generation.
\newblock In {\em Proceedings of the IEEE/CVF Conference on Computer Vision and Pattern Recognition}, pages 2002--2011, 2023.

\bibitem[\protect\citeauthoryear{Ke \bgroup \em et al.\egroup }{2024}]{ke2024repurposing}
Bingxin Ke, Anton Obukhov, Shengyu Huang, Nando Metzger, Rodrigo~Caye Daudt, and Konrad Schindler.
\newblock Repurposing diffusion-based image generators for monocular depth estimation.
\newblock In {\em Proceedings of the IEEE/CVF Conference on Computer Vision and Pattern Recognition}, pages 9492--9502, 2024.

\bibitem[\protect\citeauthoryear{Koch \bgroup \em et al.\egroup }{2019}]{ibims}
Tobias Koch, Lukas Liebel, Friedrich Fraundorfer, and Marco K{\"o}rner.
\newblock Evaluation of cnn-based single-image depth estimation methods.
\newblock In Laura Leal-Taix{\'e} and Stefan Roth, editors, {\em Computer Vision -- ECCV 2018 Workshops}, pages 331--348, Cham, 2019. Springer International Publishing.

\bibitem[\protect\citeauthoryear{Kuznetsova \bgroup \em et al.\egroup }{2020}]{kuznetsova2020open}
Alina Kuznetsova, Hassan Rom, Neil Alldrin, Jasper Uijlings, Ivan Krasin, Jordi Pont-Tuset, Shahab Kamali, Stefan Popov, Matteo Malloci, Alexander Kolesnikov, et~al.
\newblock The open images dataset v4: Unified image classification, object detection, and visual relationship detection at scale.
\newblock {\em International journal of computer vision}, 128(7):1956--1981, 2020.

\bibitem[\protect\citeauthoryear{Laina \bgroup \em et al.\egroup }{2016}]{laina2016deeper}
Iro Laina, Christian Rupprecht, Vasileios Belagiannis, Federico Tombari, and Nassir Navab.
\newblock Deeper depth prediction with fully convolutional residual networks.
\newblock In {\em 2016 Fourth international conference on 3D vision (3DV)}, pages 239--248. IEEE, 2016.

\bibitem[\protect\citeauthoryear{Li \bgroup \em et al.\egroup }{2019}]{li2019learning}
Zhengqi Li, Tali Dekel, Forrester Cole, Richard Tucker, Noah Snavely, Ce~Liu, and William~T Freeman.
\newblock Learning the depths of moving people by watching frozen people.
\newblock In {\em Proceedings of the IEEE/CVF conference on computer vision and pattern recognition}, pages 4521--4530, 2019.

\bibitem[\protect\citeauthoryear{Loshchilov and Hutter}{2019}]{loshchilov2018decoupled}
Ilya Loshchilov and Frank Hutter.
\newblock Decoupled weight decay regularization.
\newblock In {\em International Conference on Learning Representations}, 2019.

\bibitem[\protect\citeauthoryear{Melas-Kyriazi \bgroup \em et al.\egroup }{2023}]{melas2023realfusion}
Luke Melas-Kyriazi, Iro Laina, Christian Rupprecht, and Andrea Vedaldi.
\newblock Realfusion: 360deg reconstruction of any object from a single image.
\newblock In {\em Proceedings of the IEEE/CVF conference on computer vision and pattern recognition}, pages 8446--8455, 2023.

\bibitem[\protect\citeauthoryear{Ning and Gan}{2023}]{ning2023trap}
Chao Ning and Hongping Gan.
\newblock Trap attention: Monocular depth estimation with manual traps.
\newblock In {\em Proceedings of the IEEE/CVF Conference on Computer Vision and Pattern Recognition}, pages 5033--5043, 2023.

\bibitem[\protect\citeauthoryear{Patakin \bgroup \em et al.\egroup }{2022}]{patakin2022single}
Nikolay Patakin, Anna Vorontsova, Mikhail Artemyev, and Anton Konushin.
\newblock Single-stage 3d geometry-preserving depth estimation model training on dataset mixtures with uncalibrated stereo data.
\newblock In {\em Proceedings of the IEEE/CVF Conference on Computer Vision and Pattern Recognition}, pages 1705--1714, 2022.

\bibitem[\protect\citeauthoryear{Peebles and Xie}{2023}]{peebles2023scalable}
William Peebles and Saining Xie.
\newblock Scalable diffusion models with transformers.
\newblock In {\em Proceedings of the IEEE/CVF International Conference on Computer Vision}, pages 4195--4205, 2023.

\bibitem[\protect\citeauthoryear{Radford \bgroup \em et al.\egroup }{2019}]{radford2019language}
Alec Radford, Jeffrey Wu, Rewon Child, David Luan, Dario Amodei, Ilya Sutskever, et~al.
\newblock Language models are unsupervised multitask learners.
\newblock {\em OpenAI blog}, 1(8):9, 2019.

\bibitem[\protect\citeauthoryear{Ranftl \bgroup \em et al.\egroup }{2020}]{ranftl2020towards}
Ren{\'e} Ranftl, Katrin Lasinger, David Hafner, Konrad Schindler, and Vladlen Koltun.
\newblock Towards robust monocular depth estimation: Mixing datasets for zero-shot cross-dataset transfer.
\newblock {\em IEEE transactions on pattern analysis and machine intelligence}, 44(3):1623--1637, 2020.

\bibitem[\protect\citeauthoryear{Ranftl \bgroup \em et al.\egroup }{2021}]{ranftl2021vision}
Ren{\'e} Ranftl, Alexey Bochkovskiy, and Vladlen Koltun.
\newblock Vision transformers for dense prediction.
\newblock In {\em Proceedings of the IEEE/CVF international conference on computer vision}, pages 12179--12188, 2021.

\bibitem[\protect\citeauthoryear{Roberts \bgroup \em et al.\egroup }{2021}]{roberts2021hypersim}
Mike Roberts, Jason Ramapuram, Anurag Ranjan, Atulit Kumar, Miguel~Angel Bautista, Nathan Paczan, Russ Webb, and Joshua~M Susskind.
\newblock Hypersim: A photorealistic synthetic dataset for holistic indoor scene understanding.
\newblock In {\em Proceedings of the IEEE/CVF international conference on computer vision}, pages 10912--10922, 2021.

\bibitem[\protect\citeauthoryear{Rombach \bgroup \em et al.\egroup }{2022}]{rombach2022high}
Robin Rombach, Andreas Blattmann, Dominik Lorenz, Patrick Esser, and Bj{\"o}rn Ommer.
\newblock High-resolution image synthesis with latent diffusion models.
\newblock In {\em Proceedings of the IEEE/CVF conference on computer vision and pattern recognition}, pages 10684--10695, 2022.

\bibitem[\protect\citeauthoryear{Schops \bgroup \em et al.\egroup }{2017}]{schops2017multi}
Thomas Schops, Johannes~L Schonberger, Silvano Galliani, Torsten Sattler, Konrad Schindler, Marc Pollefeys, and Andreas Geiger.
\newblock A multi-view stereo benchmark with high-resolution images and multi-camera videos.
\newblock In {\em Proceedings of the IEEE conference on computer vision and pattern recognition}, pages 3260--3269, 2017.

\bibitem[\protect\citeauthoryear{Schuhmann \bgroup \em et al.\egroup }{2022}]{schuhmann2022laion}
Christoph Schuhmann, Romain Beaumont, Richard Vencu, Cade Gordon, Ross Wightman, Mehdi Cherti, Theo Coombes, Aarush Katta, Clayton Mullis, Mitchell Wortsman, et~al.
\newblock Laion-5b: An open large-scale dataset for training next generation image-text models.
\newblock {\em Advances in Neural Information Processing Systems}, 35:25278--25294, 2022.

\bibitem[\protect\citeauthoryear{Silberman \bgroup \em et al.\egroup }{2012}]{silberman2012indoor}
Nathan Silberman, Derek Hoiem, Pushmeet Kohli, and Rob Fergus.
\newblock Indoor segmentation and support inference from rgbd images.
\newblock In {\em Computer Vision--ECCV 2012: 12th European Conference on Computer Vision, Florence, Italy, October 7-13, 2012, Proceedings, Part V 12}, pages 746--760. Springer, 2012.

\bibitem[\protect\citeauthoryear{Sun \bgroup \em et al.\egroup }{2024}]{sun2024autoregressive}
Peize Sun, Yi~Jiang, Shoufa Chen, Shilong Zhang, Bingyue Peng, Ping Luo, and Zehuan Yuan.
\newblock Autoregressive model beats diffusion: Llama for scalable image generation.
\newblock {\em arXiv preprint arXiv:2406.06525}, 2024.

\bibitem[\protect\citeauthoryear{Tian \bgroup \em et al.\egroup }{2024}]{tian2024visual}
Keyu Tian, Yi~Jiang, Zehuan Yuan, Bingyue Peng, and Liwei Wang.
\newblock Visual autoregressive modeling: Scalable image generation via next-scale prediction.
\newblock {\em arXiv preprint arXiv:2404.02905}, 2024.

\bibitem[\protect\citeauthoryear{Van Den~Oord \bgroup \em et al.\egroup }{2017}]{van2017neural}
Aaron Van Den~Oord, Oriol Vinyals, et~al.
\newblock Neural discrete representation learning.
\newblock {\em Advances in neural information processing systems}, 30, 2017.

\bibitem[\protect\citeauthoryear{Wang \bgroup \em et al.\egroup }{2019}]{wang2019pseudo}
Yan Wang, Wei-Lun Chao, Divyansh Garg, Bharath Hariharan, Mark Campbell, and Kilian~Q Weinberger.
\newblock Pseudo-lidar from visual depth estimation: Bridging the gap in 3d object detection for autonomous driving.
\newblock In {\em Proceedings of the IEEE/CVF conference on computer vision and pattern recognition}, pages 8445--8453, 2019.

\bibitem[\protect\citeauthoryear{Wang \bgroup \em et al.\egroup }{2023}]{wang2023dformer}
Hefeng Wang, Jiale Cao, Rao~Muhammad Anwer, Jin Xie, Fahad~Shahbaz Khan, and Yanwei Pang.
\newblock Dformer: Diffusion-guided transformer for universal image segmentation.
\newblock {\em arXiv preprint arXiv:2306.03437}, 2023.

\bibitem[\protect\citeauthoryear{Wofk \bgroup \em et al.\egroup }{2019}]{wofk2019fastdepth}
Diana Wofk, Fangchang Ma, Tien-Ju Yang, Sertac Karaman, and Vivienne Sze.
\newblock Fastdepth: Fast monocular depth estimation on embedded systems.
\newblock In {\em 2019 International Conference on Robotics and Automation (ICRA)}, pages 6101--6108. IEEE, 2019.

\bibitem[\protect\citeauthoryear{Yang \bgroup \em et al.\egroup }{2024}]{yang2024depth}
Lihe Yang, Bingyi Kang, Zilong Huang, Xiaogang Xu, Jiashi Feng, and Hengshuang Zhao.
\newblock Depth anything: Unleashing the power of large-scale unlabeled data.
\newblock In {\em Proceedings of the IEEE/CVF Conference on Computer Vision and Pattern Recognition}, pages 10371--10381, 2024.

\bibitem[\protect\citeauthoryear{Yin \bgroup \em et al.\egroup }{2021}]{yin2021learning}
Wei Yin, Jianming Zhang, Oliver Wang, Simon Niklaus, Long Mai, Simon Chen, and Chunhua Shen.
\newblock Learning to recover 3d scene shape from a single image.
\newblock In {\em Proceedings of the IEEE/CVF Conference on Computer Vision and Pattern Recognition}, pages 204--213, 2021.

\bibitem[\protect\citeauthoryear{Yin \bgroup \em et al.\egroup }{2023}]{yin2023metric3d}
Wei Yin, Chi Zhang, Hao Chen, Zhipeng Cai, Gang Yu, Kaixuan Wang, Xiaozhi Chen, and Chunhua Shen.
\newblock Metric3d: Towards zero-shot metric 3d prediction from a single image.
\newblock In {\em Proceedings of the IEEE/CVF International Conference on Computer Vision}, pages 9043--9053, 2023.

\bibitem[\protect\citeauthoryear{Yu \bgroup \em et al.\egroup }{2021}]{yu2021vector}
Jiahui Yu, Xin Li, Jing~Yu Koh, Han Zhang, Ruoming Pang, James Qin, Alexander Ku, Yuanzhong Xu, Jason Baldridge, and Yonghui Wu.
\newblock Vector-quantized image modeling with improved vqgan.
\newblock {\em arXiv preprint arXiv:2110.04627}, 2021.

\bibitem[\protect\citeauthoryear{Yu \bgroup \em et al.\egroup }{2023}]{yu2023language}
Lijun Yu, Jos{\'e} Lezama, Nitesh~B Gundavarapu, Luca Versari, Kihyuk Sohn, David Minnen, Yong Cheng, Agrim Gupta, Xiuye Gu, Alexander~G Hauptmann, et~al.
\newblock Language model beats diffusion--tokenizer is key to visual generation.
\newblock {\em arXiv preprint arXiv:2310.05737}, 2023.

\end{thebibliography}
